\documentclass[journal]{IEEEtran}

\ifCLASSINFOpdf
\else
\fi

\usepackage{amsmath,amsfonts,amssymb}
\usepackage{color,soul}
\usepackage{epstopdf}
\usepackage{tikz}
\usepackage{balance}
\usepackage{url}
\usepackage{multirow,hhline}
\usepackage{placeins}
\usepackage{wasysym}
\usepackage[nomessages]{fp}

\newcommand{\tabPath}{./R1.tables/}
\newcommand{\figPath}{./R1.figures/}
\newcommand{\AdeImFolder}{\figPath samples_selected/Adelaide/}
\newcommand{\StockImFolder}{\figPath samples_selected/Stockholm/}
\newcommand{\WashImFolder}{\figPath samples_selected/Washington/}

\newcommand{\Dlambda}{{\bf D}$_\lambda$}
\newcommand{\Desse}{{\bf D}$_S$}
\newcommand{\ru}{\rule{0mm}{3.2mm}}

\newcommand{\bx}{\mathbf{x}}
\newcommand{\by}{\mathbf{y}}
\newcommand{\bw}{\mathbf{w}}
\newcommand{\bz}{\mathbf{z}}
\newcommand{\bbb}{\mathbf{b}}

\newcommand{\image}{\pgfuseimage}

\pgfdeclareimage[width=0.95\columnwidth]{residuo}{\figPath residual.pdf}
\pgfdeclareimage[width=\columnwidth]{net}{\figPath net.pdf}

\pgfdeclareimage[width=0.9\columnwidth]{sam}{\figPath sam.pdf}
\pgfdeclareimage[width=0.9\columnwidth]{sid}{\figPath sid.pdf}
\pgfdeclareimage[width=0.9\columnwidth]{L2}{\figPath L2.pdf}
\pgfdeclareimage[width=0.9\columnwidth]{L1}{\figPath L1.pdf}
\pgfdeclareimage[width=\columnwidth]{Consistency}{\figPath Consistency.pdf}
\pgfdeclareimage[width=0.95\columnwidth]{up}{\figPath up.new.pdf}
\pgfdeclareimage[width=1.6\columnwidth]{finetune}{\figPath finetune.pdf}

\pgfdeclareimage[width=0.64\columnwidth]{Q_Caserta}{\figPath Q_Caserta.pdf}
\pgfdeclareimage[width=0.64\columnwidth]{Q_Ade}{\figPath Q_Ade.pdf}
\pgfdeclareimage[width=0.64\columnwidth]{Q_Wash}{\figPath Q_Wash.pdf}

\pgfdeclareimage{TabAde}{\tabPath WV3_Adelaide.pdf}
\pgfdeclareimage{TabStock}{\tabPath WV2_Stockholm.pdf}
\pgfdeclareimage{TabWash}{\tabPath WV2_Washington.pdf}
\pgfdeclareimage{TabWV}{\tabPath WV.pdf}
\pgfdeclareimage{TabGE}{\tabPath GE.pdf}
\pgfdeclareimage{TabIK}{\tabPath IK.pdf}

\pgfdeclareimage{TabCplx}{\tabPath complexityFull.pdf}

\pgfdeclareimage{T6Stock}{\StockImFolder T6Stockholm.pdf}
\pgfdeclareimage{T5StockC}{\StockImFolder T5Stockholm_conf.pdf}
\pgfdeclareimage{T60Wash}{\WashImFolder T60Washington.pdf}
\pgfdeclareimage{T3WashC}{\WashImFolder T3Washington_conf_resized.pdf}
\pgfdeclareimage{T20bAdeC}{\AdeImFolder T20bAdelaide_conf_resized.pdf}

\definecolor{colPeppe}{rgb}{0.7, 0.2, 0.15}

\definecolor{colWarning}{rgb}{0.7, 0.2, 0.7}

\begin{document}

\title{Target-adaptive CNN-based pansharpening}

\author{
Giuseppe~Scarpa, 
Sergio~Vitale,
and~Davide~Cozzolino
\thanks{G. Scarpa and D. Cozzolino are with the Department of Electrical Engineering and Information Technology,
University Federico II, Naples, Italy, e-mail: \{firstname.lastname\}@unina.it.
S. Vitale is with the Engineering Department, University Parthenope, Naples, Italy,
e-mail: sergio.vitale@uniparthenope.it.}
}

\maketitle

\begin{abstract}
We recently proposed a convolutional neural network (CNN) for remote sensing image pansharpening
obtaining a significant performance gain over the state of the art.
In this paper, we explore a number of architectural and training variations to this baseline,
achieving further performance gains with a lightweight network which trains very fast.
Leveraging on this latter property, we propose a target-adaptive usage modality
which ensures a very good performance also in the presence of a mismatch w.r.t. the training set, and even across different sensors.
The proposed method, published online as an off-the-shelf software tool,
allows users to perform fast and high-quality CNN-based pansharpening of their own target images on general-purpose hardware.
\end{abstract}

%

\IEEEpeerreviewmaketitle

\section{Introduction}

Thanks to continuous technological progresses,
there has been a steady improvement in the quality of remote sensing products, and especially in the spatial and spectral resolution of images.
Then, when technology reaches its limits, signal processing methods may provide a further quality boost.
Pansharpening is among the most successful examples of such a phenomenon.
Given a high spatial resolution panchromatic band (PAN) and a low resolution multispectral stack (MS),
it generates a datacube at the highest resolution in both the spectral and spatial domains.
Results are already promising,
but intense research is going on to approach more and more closely the quality of
ideal 
high-resolution data.

In the last decades, many different approaches have been proposed to address the pansharpening problem.
A classic approach is the component substitution (CS) \cite{Shettigara1992},
where the multispectral component is upsampled and transformed in a suitable domain
and the panchromatic band is used to replace one of the transformed bands before inverse transform to the original domain.
Under the restriction that only three bands are concerned,
the Intensity-Hue-Saturation (IHS) transform can be used, with the intensity component replaced by the panchromatic band \cite{Tu2001}.
This same approach has been generalized in \cite{Tu2004} (GIHS) to include additional bands.
Many other transforms have been considered for CS like, for example,
the principal component analysis \cite{Chavez1989}, the Brovey transform \cite{Gillespie1987}, and Gram-Schmidt (GS) decomposition \cite{Laben2000}.
More recently,
{\em Adaptive} CS methods have also been introduced,
like the advanced versions of GIHS and GS adopted in \cite{Aiazzi2007}, the partial substitution method (PRACS) proposed in \cite{Choi2011},
and optimization-based techniques \cite{Garzelli2008}.

CS methods approach the pansharpening problem from the spectral perspective, as the fusion occurs with respect to a spectral transform.
Another class of methods regard the problem from the geometric, or spatial, perspective
mostly relying on multiresolution analysis (MRA) \cite{Ranchin2000}.
MRA-based methods aim to extract spatial details from the PAN component to be later injected in the resampled MS component.
Spatial details can be extracted in different ways,
using, for example, decimated or undecimated wavelet transforms \cite{Nunez1999, Ranchin2000, Otazu2005, Khan2008},
Laplacian pyramids \cite{Aiazzi2002, Aiazzi2003, Aiazzi2006, Lee2010, Restaino2017},
or other nonseparable transforms, like the contourlet \cite{Shah2008}.

The separation between CS and MRA methods is neither always clear-cut nor exhaustive.
There are in fact many examples of methods which are better cast as statistical
\cite{Fasbender2008, Zhang2012, Garzelli2015, Meng2015, Shen2016, Zhong2017}.
or variational
\cite{Palsson2014, Duran2017}
that get state of the art results.
However this CS-MRA dichotomy is useful to understand the behaviour of any method falling
in these categories as highlighted in \cite{Vivone2015, Aiazzi2017}.
Specifically,
given well registered MS-PAN components, and accurate modeling of the sensor Modulation Transfer Function (MTF),
methods based on MRA achieve usually a better pansharpening quality than those based on CS \cite{Alparone2016}.
On the contrary,
when MS-PAN misregistration occurs, both CS and MRA methods may lose geometric sharpness,
but the latter suffer also from spectral mismatch,
making them unsuitable for applications where spectral accuracy is of critical importance \cite{Aiazzi2017}.

In the last few years machine learning methods have gained much attention from both signal processing and remote sensing communities.
Compressive sensing and dictionary based methods,
for example,
have been successfully applied to pansharpening in
several papers \cite{Li2011, Li2013, Zhu2013, Cheng2014, Vicinanza2015, Zhu2016}.
Very recently,
following the recent technological and theoretical advances in computer vision and related fields,
also {\em deep} learning methods have been applied
to remote sensing problems \cite{Castelluccio2015,Ienco2017,Zhu2017},
and several papers have been proposed to address pansharpening \cite{Huang2015, Masi2016,Wei2017a, Wei2017, Rao2017, Azarang2017}.
In particular, our CNN-based solution \cite{Masi2016}, inspired by work on super-resolution \cite{Dong2016},
provided a significant performance improvement with respect to the previous state-of-the-art.

\begin{figure}[t]
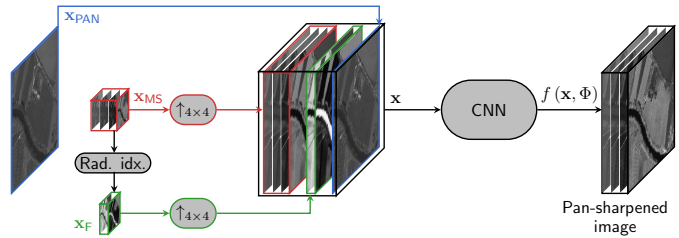

	\centering \image{net}
	\caption{General workflow of the PNN algorithm \cite{Masi2016}.}
	\label{fig:baseline}
\end{figure}

In this work we start from the baseline solution of \cite{Masi2016}
and explore a number of variations aimed at improving both performance and robustness,
including different learning strategies, cost functions, and architectural choices.
The most remarkable improvements are obtained by including a target-adaptive tuning phase,
which solves to a large extent the problem of insufficient training data,
allowing users to apply the proposed architecture to their own data and achieve good results consistently.
The proposed solutions are extensively tested on images acquired by a number of sensors,
covering different spatial and spectral resolutions.
A substantial improvement is observed over both the baseline and the state-of-the-art methods
available in the {\em Open Remote Sensing} repository \cite{OpenRS},
under a wide range of quality measures.

In the rest of the paper we describe
the baseline method (Section II),
the proposed architectural improvements (Section III),
the target adaptive solution (Section IV), and
the experimental results (Section V),
before drawing conclusions (Section VI).

\section{A pansharpening Neural Network}
\label{sec:related}

In \cite{Masi2016, Masi2017}, inspired to previous work on super-resolution \cite{Dong2016}, we proposed the Pansharpening Neural Network (PNN), summarized in the block diagram of Fig.\ref{fig:baseline}.
The core of the network is a simple three-layer convolutional neural network (CNN), not shown here for brevity.
The CNN takes in input
the panchromatic band $\bx_{\rm PAN}$ (blue),
the multispectral component $\bx_{\rm MS}$ (red) up-sampled via polynomial interpolation,
and a few radiometric indices $\bx_{\rm F}$ (green) extracted from the MS component and interpolated as well.
The latter component, comprising some nonlinear combinations of MS spectral bands, has proven experimentally to improve the network performance.

The CNN is composed by three convolutional layers
with nonlinear activations in both the input and the hidden layers, and linear activation in the output layer.
For each layer, assuming $N$ input bands, $M$ output bands, and filters with $K\times K$ receptive field (spatial support),
a number of parameters must be learned on the training set,
a $M\times [N \times (K\times K)]$ tensor, $\bw$, accounting for the weights, and a $M$-vector, $\bbb$, for the biases.
For layer $l$, with input $\bx^{(l)}$, the filter output is computed as
\[
    \bz^{(l)} = \bw^{(l)} \ast \bx^{(l)} + \bbb^{(l)},
\]
where the $m$-th component can be expressed as
\[
\bz^{(l)}(m,\cdot,\cdot) = \sum_{n=1}^N  \bw^{(l)}(m,n,\cdot,\cdot) \ast \bx^{(l)}(n,\cdot,\cdot)+ \bbb^{(l)}(m)
\]
in terms of the usual 2D convolution.
After filtering, in the input and the hidden layers
a pointwise nonlinear function is applied, in particular a Rectified Linear Unit, ReLU$(\cdot)=\max(0,~\cdot)$,
to obtain the actual feature maps $\by^{(l)}$
\begin{equation}
    \by^{(l)} =f_l(\bx^{(l)},\Phi_l) = \max(0,\bz^{(l)})
    \label{eq:activation}
\end{equation}
where $\Phi_l\triangleq\left(\bw^{(l)},\bbb^{(l)}\right)$.
The choice of ReLU nonlinearities is motivated experimentally \cite{Krizhevsky2012} by the good convergence properties they guarantee.
Notice that, as neither stride nor pooling are used, each layer preserves the input resolution
and hence $\bx^{(l)}$ and $\by^{(l)}$ have the same spatial size.

\subsection{Learning}
\label{ssec:learning}

Let $\bx = (\bx_{\rm PAN}, \bx_{\rm MS}, \bx_{\rm F})$ denote the composite CNN input stack
and $f(\bx,\Phi)$ be the overall function computed by the CNN,
with $\Phi=\left(\Phi_1,\Phi_2,\Phi_3\right)$ the collection of its parameters.
In order to learn the network parameters,
reference data are required, that is, examples of perfectly pansharpened images coming from the same sensor for which the network is designed.
Unfortunately, images at full spatial and spectral resolution are not available,
which complicates training and performance assessment alike.

\begin{figure}
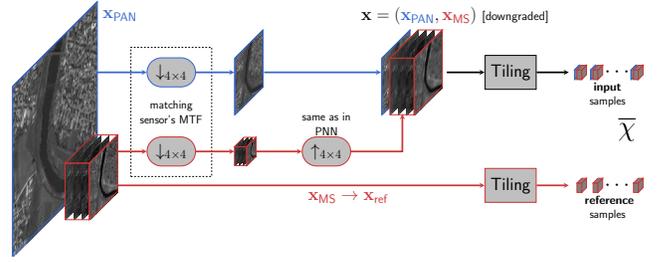

	\centering \image{up}
	\caption{Generation of a training dataset through Wald's protocol.}
	\label{fig:wald}
\end{figure}

For the purpose of validation, this problem is often addressed by resorting to the Wald protocol \cite{Wald97},
which consists in downsampling both the PAN and the MS components,
so that the original MS component can be taken as a reference for the pansharpening of the downsampled data.
Fig.~\ref{fig:wald} shows how Wald's protocol is used to create training examples.
Before downsampling, a low-pass filter is applied to reduce aliasing.
To match the sensor properties, in \cite{Aiazzi2006} it is proposed to use an approximation of the sensor 
MTF.
This scheme generalizes readily to the case in which additional low-resolution input bands are considered,
like the MS radiometric indices.

Given a training set $\overline{\chi}=\left\{ \overline{\bx}_1,\ldots, \overline{\bx}_Q \right\}$,
generated by Wald's protocol with any MTF modeling,
comprising $Q$ input-output image pairs $\overline{\bx} \triangleq \left(\bx,\bx_{\rm ref}\right)$,
the objective of the training phase is to find
\begin{equation}
    \Phi =          \underset{\Phi}{\arg\min} \; J\left(\overline{\chi},\Phi\right)
		 \triangleq \underset{\Phi}{\arg\min} \; \frac{1}{Q}\sum_{\overline{\bx} \in \overline{\chi}} L(\overline{\bx},\Phi) \nonumber
\end{equation}
where $L(\overline{\bx},\Phi)$ is a suitable loss function.

In \cite{Masi2016} the mean square error (MSE) was used as loss function\footnote{To account for border effects,
the norm is computed on cropped versions of $f(\bx,\Phi)$ and $\bx_{\rm ref}$.}
\[
    L(\overline{\bx},\Phi) \propto || f(\bx,\Phi) - \bx_{\rm ref}  ||^2_2
\]
and the minimization was carried out by stochastic gradient descent (SGD) with momentum \cite{Sutskever2013}.
In particular,
the training set was partitioned in $P$ batches, $\left\{B_1,\ldots,B_{P} \right\}$, with $\bigcup_j B_j = \overline{\chi}$,
and at each iteration a batch was sampled and used to estimate the gradient and update parameters as
\begin{eqnarray}
    &&  \nu^{(n+1)} \leftarrow \mu \nu^{(n)}  + \alpha \nabla_{\Phi} J\left(B_{j_n},\Phi^{(n)} \right) \nonumber \\
    && \Phi^{(n+1)} \leftarrow \Phi^{(n)}  - \nu^{(n+1)} \nonumber
\end{eqnarray}
Training efficiency and accuracy depend heavily on the algorithm hyperparameters,
learning rate, $\alpha$,  momentum, $\mu$, and velocity, $\nu$,
the most critical being the learning rate, which can cause instability when too large or slow convergence when too small.
In \cite{Masi2016}, after extensive experiments, the learning rate was set to $10^{-4}$ for $\Phi_1$ and $\Phi_2$ and to $10^{-5}$ for $\Phi_3$,
which ensured convergence in about $10^6$ iterations.

\section{Improving CNN-based pansharpening}
\label{sec:proposed}

Although the PNN architecture relies on solid conceptual foundations,
following a well-motivated path from dictionary-based super-resolution \cite{Yang2010}, to its CNN-based counterpart \cite{Dong2016}, and finally to pansharpening,
there is plenty of room for variations and, possibly, further improvements.
Therefore we explored experimentally a number of alternative architectures and learning modalities.
In the following, we describe only the choices that led to significant improvements or are otherwise worth analyzing,
that is,
\begin{itemize}
\item   using L1 loss;
\item   working on image residuals;
\item   using deeper architectures.
\end{itemize}

\subsection{Using L1 loss function}

\begin{figure}
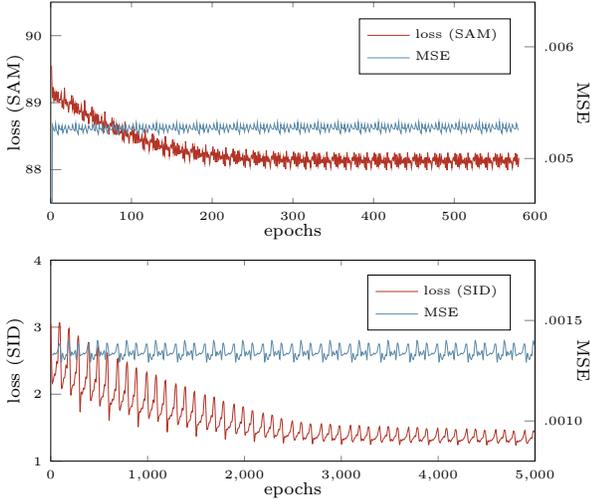

\centering 
\image{sam}

\image{sid}
\caption{Loss and MSE during training using 
SAM (top) or SID (bottom) as loss.}
\label{fig:sam}
\end{figure}

In deep learning, choosing the ``right'' loss function can make the difference between being stuck with disappointing results and achieving the desired output.
With pansharpening there is no shortage of candidate loss functions \cite{Vivone2015,Masi2016},
but some of them are quite complicated, hence time-consuming, and may happen to be unstable.
As an example one may think to use typical measures employed
for quality assessment or discrimination 
in multispectral or hyperspectral image analysis, 
like the spectral angle mapper (SAM) \cite{Yuhas1992}
or the spectral information divergence (SID) \cite{Chang2000}.
Unless some form of regularization is considered,
these choices are unsuited for optimization 
as they are insensitive to intensity scaling
and because they induce highly complex targets,
full of local minima, 
where one can easily get trapped during training.
This can also be observed experimentally with the help
of Fig.~\ref{fig:sam}, where we plot the loss (SAM or SID) 
along with the a common measure of distortion, the 
mean square error (MSE), as function of the traininig iterations.
As it can be seen, the losses
do not reach reasonable levels of accuracy,
getting trapped in some local minima.
Infact, typical values should be a few degrees, for SAM,
and some order of magnitude lesser than the unit,
in case of SID.
Furthermore, as we will see later,
for our dataset we expect the MSE to be one order of magnitude smaller than the values registered in Fig.~\ref{fig:sam}.
More in general,
the MSE do not follow the trend of the loss,
which is an indirect confirm of the scaling insensity of the selected distances.

Motivated by the above observations
we decided to restrict our study to Ln norms,
as the reduction of the training time is 
a primary goal in this work,
leaving to future research
the aim of improving this choice further.
In \cite{Masi2016} we used the L2 loss, as in \cite{Dong2016}, but our preliminary experiments proved the L1 norm to be a much better choice.
Surprisingly, by training the network to minimize a L1 loss, we achieved better results even in terms of MSE or other L2 related indicators.
This behaviour is in general possible because of the non convexity of the target and,
indeed, has already been observed and discussed in the deep learning literature \cite{Girshick2015}.
On one hand, when the regression targets are unbounded, training with L2 loss requires careful tuning of learning rates in order to prevent exploding gradients.
On the other hand, and probably more important, the L2 norm penalizes heavily large errors but is less sensitive to small errors,
which means that the learning process slows down significantly as the output approaches the objective.
This is the working point of highest interest for our application,
because the quality of CNN-based pansharpening is already very good, according to both numerical indicators and visual inspection \cite{Masi2016}.
To achieve further improvements, one must focus on small errors, a goal for which the L1 norm is certainly more fit.
On the down side, the L1 norm is more prone to instabilities
but in our experiments these never prevented eventual convergence and satisfactory results.

\begin{figure}
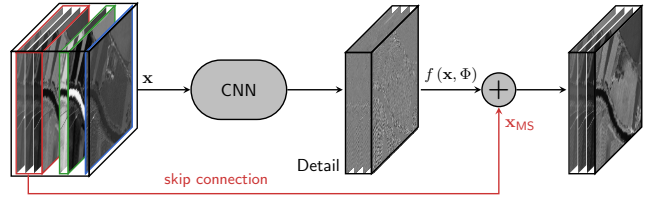

\centering \image{residuo}
\caption{Proposed residual-based architecture. Preprocessing not shown.}
\label{fig:residuo}
\end{figure}

\begin{figure*}
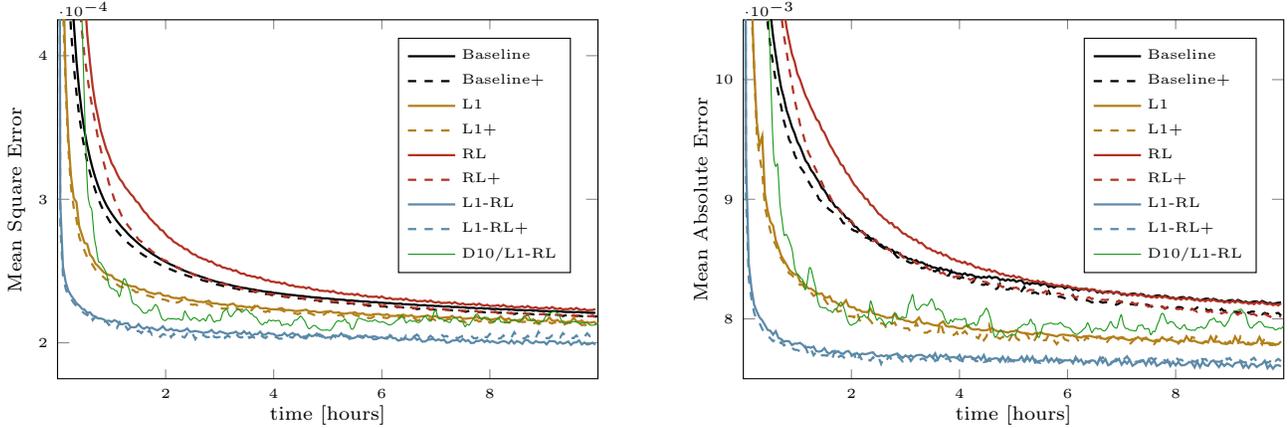

\image{L2} \hspace{9mm} \image{L1}
\caption{Mean square error (MSE, left) and mean absolute error (MAE, right) of the proposed methods on the GeoEye-1 validation set.
Dashed lines (+ in the legend) indicate input augmentation with radiometric indices.
Accordingly, baseline+ coincides with PNN \cite{Masi2016}.
L1 indicates use of L1-loss in place of L2-loss for training.
RL indicates use of residual learning.
D10 marks the best deep architecture (10 layers) selected by preliminary tests.}
\label{fig:Preliminary}
\end{figure*}

\subsection{Working on image residuals}

In the baseline architecture proposed in \cite{Masi2016}, the network is trained to reconstruct the whole target image.
However, the low-pass component of the output, that is, the up-sampled MS component, is already available and only the high-resolution residual need be generated.
Based on this observation, we modified the network to let it reconstruct only the missing part of the desired output.

The residual-based version of our baseline solution is shown in Fig.\ref{fig:residuo},
where the preprocessing is omitted for the sake of simplicity.
The core CNN is trained to generate only the residual component, namely the desired pansharpened image minus its low-pass component.
Therefore, the desired output is obtained by summing the up-sampled MS component, $\bx_{\rm MS}$, made available through a skip connection, to the network output, $f\left(\bx,\Phi\right)$.
The loss is then computed as
\begin{eqnarray}
    L(\overline{\bx},\Phi) & \propto & || (f(\bx,\Phi) + \bx_{\rm MS}) - \bx_{\rm ref} ||^2_2 \nonumber \\
		                   & \propto & ||  f(\bx,\Phi) - \Delta \bx_{\rm ref} ||^2_2 \nonumber
\end{eqnarray}
with the residual reference defined as $\Delta \bx_{\rm ref} \triangleq \bx_{\rm ref} - \bx_{\rm MS}$ and, accordingly,
the training samples redifined as $\overline{\bx} \triangleq (\bx,\Delta\bx_{\rm ref})$.
From the training perspective, the only difference with the baseline solution of Fig.\ref{fig:baseline} is that reference data are obtained by subtracting the interpolated MS component $\bx_{\rm MS}$.

If we also replace the L2 norm with L1 norm as suggested in the previous subsection, the loss becomes eventually
\[
		L(\overline{\bx},\Phi) \propto || f(\bx,\Phi) - \Delta \bx_{\rm ref}  ||_1
\]

Note that residual learning is not a new idea.
In \cite{Zeyde2012,Timofte2013} it was used for dictionary-based super-resolution, proving effective both in terms of accuracy and training speed.
More recently, it was advocated for training very deep CNNs \cite{Kim2016,He2016},
and used successfully in various applicative fields, e.g. image denoising with very deep networks \cite{Zhang2017}.
Residuals were also used for pansharpening in \cite{Vicinanza2015}, in the context of sparse representation,
and in \cite{Wang2016}, where a residual-based regressor was proposed.
As said before, residual learning is a natural choice for pansharpening, due to the availability of the low-pass component.
More in general, it was observed experimentally \cite{He2016} that
training the network to reproduce the desired output may be quite difficult when the output is itself very similar to the input.
The process becomes much more efficient when targeting {\em differences} between input and output, that is, residuals.
Therefore this applies to many image restoration and enhancement tasks, such as denosing, super-resolution, and pansharpening.

Very recently,
two groups of researchers have proposed using residual learning for pansharpening \cite{Rao2017, Wei2017a, Wei2017},
in the context of deep or very deep CNNs, claiming some performance improvements w.r.t. PNN.
However, in both cases the experimental validation is somewhat faulty, preventing solid conclusions.
In \cite{Rao2017} experiments are carried out on Landsat 7 images,
with geometric and spectral characteristics very far from those of typical multiresolution images of interest.
In \cite{Wei2017a,Wei2017}, instead, the assessment involves only reduced resolution data.
Therefore, in both cases there is no clue on how the methods perform on full-resolution images of interest.

\subsection{Using deeper architectures}

A generic CNN is formed by the cascade of $L$ processing layers, hence if computes a composite function in the form
\begin{equation}
    f(\bx,\Phi) = f_L(f_{L-1}(\ldots f_1(\bx,\Phi_1),\ldots,\Phi_{L-1}),\Phi_L)
    \label{eq:chain}
\end{equation}
where $\Phi\triangleq(\Phi_1,\ldots,\Phi_L)$.
Our baseline method, as well as variations considered thus far, relies on a 3-layer CNN, which can be considered a rather shallow network.
On the contrary, the current trend in the literature is to use deep or very deep networks.
In principle, deeper networks exhibit a superior expressiveness, because more and more abstract features can be built on top of simpler ones.
Moreover, it has been demonstrated \cite{Goodfellow2016} that the representational capability of a network grows with its dimension.
On the downside, training very deep networks may require a long time and convergence is more difficult,
because information does not backpropagate easily through so many layers.
A number of approaches have been proposed to deal with this problem,
including residual learning, suitable losses and activation functions, a careful choice of hyper-parameters, and batch normalization.

Hence, we decided to test deeper CNNs for pansharpening.
Following the approach of \cite{Simonyan2014},
we consider $L$ identical layers, except for the input and output layers which are modified to account for the input and output shape.
Filter supports are reduced to obtain composite receptive fields that have approximately the same size as in the baseline.
Like in the baseline, we use ReLU activations for the input and the hidden layers, and an identity mapping in the output layer.
We also include already residual learning and L1 loss in the new solution.
Finally, during training, we stabilize the layers' inputs by means of batch normalization \cite{Ioffe2015}, thus removing unwanted random fluctuations and speeding-up the training phase significantly.

\begin{figure}
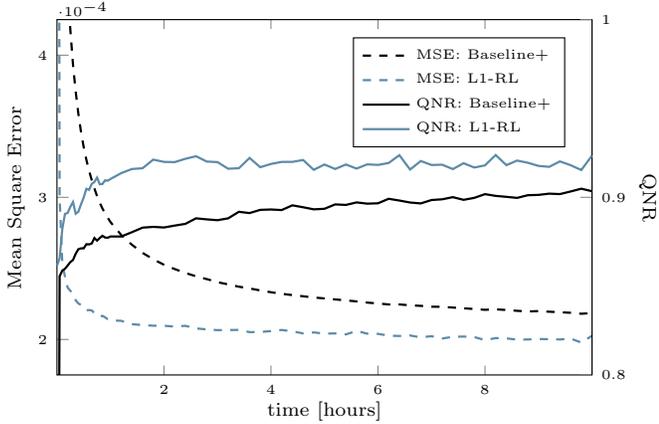

\image{Consistency}
\caption{Testing the coherence between reduced-resolution (MSE) and full-resolution (QNR) quality indicators.}
\label{fig:qnr}
\end{figure}

\subsection{Preliminary experiments}

To gain insight into the impact of the proposed improvements
we carried out some preliminary experiments on our GeoEye-1 multiresolution dataset, described in detail in Section~\ref{sec:exp}.
Performance is assessed in terms of average error on the validation dataset vs. training time.
We do not use number of iterations or epochs, as their time cost varies as a function of the architecture.
Both the mean square error (MSE) and the mean absolute error (MAE), are considered, left and right parts of Fig.~\ref{fig:Preliminary},
irrespective of the loss function, L2 or L1, used to train the CNN.
Indeed, since there is no consensus on the ideal performance measure for pansharpening,
results with two different and well-understood norms may provide some indications on robustness across other more complicated measures.

\begin{figure*}
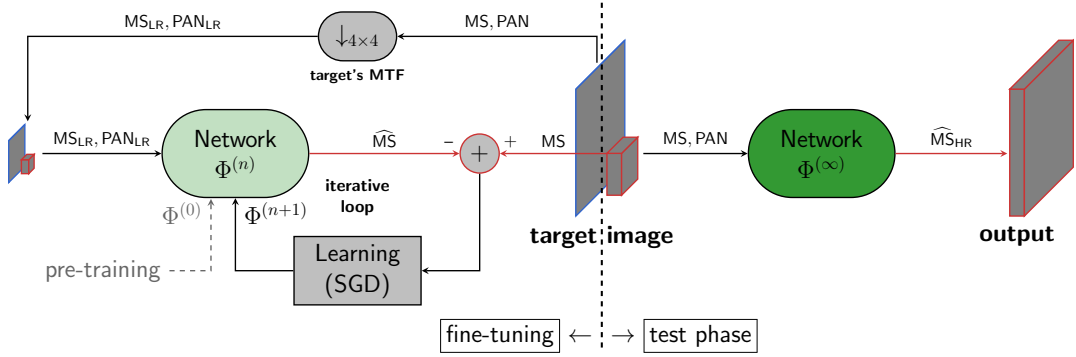

\centering
\image{finetune}
\caption{Workflow of the proposed target-adaptive modality.
The target image at center figure feeds the fine-tuning process on the left-hand side until convergence.
Then, when fine-tuning is over, it enters the fully trained network on the right-hand side to produce the desired pansharpened output.
}
\label{fig:finetune}
\end{figure*}

The main phenomena are quite clear, and consistent for the two cases.
First of all, replacing L2 with L1 norm in the training phase provides a significant performance gain.
Then, a further improvement is obtained by adopting also residual learning in combination with the L1 loss.
On the contrary, residual learning has a negative impact on performance when used in combination with L2 loss (the RL curves).
A possible explanation is that residual learning works on smaller inputs,
not well discriminated by the L2 norm, causing a slower backpropagation of errors.
Also,
increasing the network depth (always with residual learning and L1 loss) does not seem to provide any benefit,
as clear by the D10 curves, associated with the best architecture found by varying number of layers, filter support, and number of features per layer.
Despite batch normalization, and a careful setting of learning rates,
instabilities occur, and no gain is observed at convergence over the 3-layer net.
Finally, augmentation through radiometric indices,
which was found beneficial for the baseline, proves useless for the best methods emerging from this analysis.

Interesting results emerge also in terms of training speed.
Indeed, the baseline requires a long time to reach convergence, and actually the loss keeps decreasing even after 10 hours.
On the contrary, the residual-L1 version achieves the same performance after just 30 minutes,
and the training appears to be complete after 2 hours.
Therefore, besides providing a large performance gain, the new solution cuts training times by a factor 5 or more.

Note that similar results, not shown here for brevity, have been obtained with other datasets and sensors.
We also tested the coherence between full-reference quality indicators, used on downsampled data, and no-reference quality indicators, used on full-resolution data.
As an example, for two different architectures,
Fig.~\ref{fig:qnr} shows the behavior of MSE and QNR (a full-resolution quality index) as training proceeds on the GeoEye-1 dataset.
On a rough scale, results are consistent, with QNR approaching 1 (best quality) as the MSE reduces.
However, even with stable MSE, significant fluctuations in QNR are observed, maybe due to the imperfect MTF modeling,
which suggest due care when considering these measures.

In conclusion, these preliminary results allow us to proceed safely with design choices.
Specifically, from now on we will focus on the residual learning architecture of Fig.\ref{fig:residuo}, with L1 loss and without input augmentation.
Moreover, we will keep the original 3-layer CNN, as current experiments do not support
the adoption of a deeper architecture\footnote{This result,
in partial disagreement with the current literature, may be due to a number of reasons, including insufficient training data.
Therefore, we will keep considering this option in future work, and investigate it further.}.

\section{Target-adaptive pansharpening}

A basic prescription of deep learning is to train the network of interest on a large and varied dataset,
representative of the data that will be processed in actual operations.
This allows the network to {\em generalize} and provide a good performance also on data never seen during training.
On the contrary, if the training set is too small or not varied enough,
the network may {\em overfit} these data, providing a very good performance on them, and working poorly on new data.
In other words,
the network is desired to be robust over a wide distribution of data, although not optimal for any of them.
Once the training is over,
all parameters are freezed and the network is used on the targets with no further changes.
This procedure is motivated by the desire to obtain a stable and predictable network and, not least, by computational issues,
since training a deep network anew for each target would be a computational nightmare.
However, what if such a dedicated training were feasible in real time?

We explored this opportunity, and verified that including a target-adaptive fine-tuning step in our method is computationally feasible,
actually, almost transparent to the user if suitable hardware is available,
and may provide huge improvements in performance whenever a mismatch occurs between training data and target image.
Therefore, we propose, here, a target-adaptive version of our pansharpening network.
We use the best architecture emerging from the analyses of previous Section, a three-layer CNN with residual learning and L1 loss,
and train it on the available dataset.
Then, at run-time, a fine-tuning step is performed on the target data, so as to provide the desired adaptation.

The whole process is described in more detail in Fig.~\ref{fig:finetune}.
Initially, the network, represented by the light green oval\footnote{For simplicity
{\em Network} includes also the MS upsampling (refer to Fig.~\ref{fig:baseline}).}
in the left part of the figure, uses the parameters, $\Phi^{(0)}$, selected in the pre-training phase.
Then, these are iteratively fine-tuned, $\ldots, \Phi^{(n)}, \Phi^{(n+1)}, \ldots$,
by training over data sampled from the target image itself, always using Wald's protocol (note that red lines convey only MS data).
Upon convergence, the parameters are freezed to $\Phi^{(\infty)}$ and used in the final network, shown on the right as a dark green oval,
to carry out actual pansharpening of the target image.
Our experiments show that 50 iterations ensure a good quality fine-tuning in all cases,
hence, to keep complexity under control, we use this default value rather than thresholding over the loss variations.

\begin{figure*}
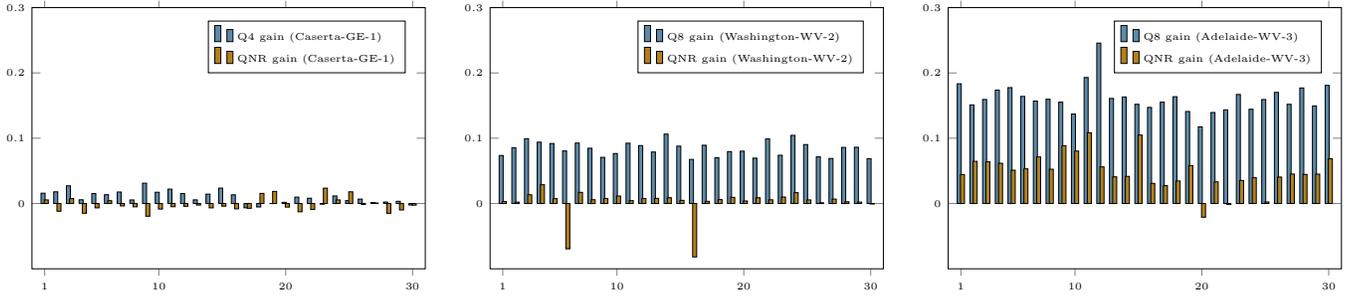

\begin{tabular}{ccc}
\image{Q_Caserta} & \image{Q_Wash} & \image{Q_Ade}
\end{tabular}
\caption{Performance gain ensured by fine-tuning over 30 target clips.
Q4/Q8: full-reference measure computed on reduced resolution data.
QNR: no-reference measure computed on full resolution data.
From left to right: favourable case (same sensor, same scene), typical case (same sensor, different scenes), challenging case (different sensors).}
\label{fig:RefGain}
\end{figure*}

It is worth emphasizing the importance of such an adaptation step for pansharpening.
Indeed, contrary to what happens in other fields,
no large database is available to the remote sensing community for developing and testing new solutions,
and one must resort to proprietary datasets, often not large and diverse as necessary.
Hence, the performance of CNN-based methods on new data may happen to be much worse than expected, even worse than conventional methods,
leaving the huge potential of deep learning untapped.

Also, it should be realized that this fine-tuning step is computationally light, unlike what happens with conventional training from scratch.
In our experiments, adapting the proposed network to a 1280$\times$1280 (PAN resolution) image
took just 1.5 seconds on a GPU-equipped computer, and about 210 seconds on a general purpose CPU,
which is appreciable by the user but certainly affordable in view of the ensuing performance improvement.
Note also that the computational overhead for fine-tuning does not grow with the image size,
as only a certain number of samples need be extracted from the target image to ensure a good adaptation.
Hence, for large real-world images, it becomes negligible w.r.t. actual pansharpening even in the CPU-only case.

This very short processing time is readily explained, in fact
\begin{itemize}
\item   the network is already pre-trained, and therefore fine-tuning requires a small number of iterations;
\item   the adaptation involves only the target data (possibly, only a subset of them) which are orders of magnitude less than the training data;
\item   the selected architecture, with residual learning, trains much faster than conventional CNNs, like for example the baseline PNN.
\end{itemize}
We also underline that the fine-tuning phase does not require any active user involvement.

Also in this case, before turning to extensive experiments,
we present some preliminary evidence of the achievable performance improvements in the various operating conditions of interest.
In particular, in Fig.~\ref{fig:RefGain} we report the performance gain over the proposed L1-RL architecture
observed on a number of different target clips when fine-tuning is performed.
Performance is measured both by the Q4/Q8 full-reference measure (see Section~\ref{sec:exp}) computed on the reduced-resolution data, and by the no-reference QNR measure mentioned before.

In the first graph, we consider a rather favourable case in which the target clips, although disjoint from the training set, are drawn from the same large image (Caserta-GE-1).
Obviously, the performance gain is very limited on all clips, due to the perfect alignment between training and test data.
Note also that, while there is always a gain for the full-reference measure Q4, this is not the case for the QNR,
underlining again the mismatch between these two classes of measures. 
In the second graph, we consider the more typical case in which training set (Caserta-WV-2) and target clips (Washington-WV-2)
concern different images taken by the same sensor.
Here, fine-tuning guarantees a significant improvement of the Q8 indicator (0.1, on the average, on a unitary scale)
confirming the potential benefit of this processing step.
Again, the objective Q8 measure is only mildly correlated with the QNR which, in some cases, exhibits even a significant drop.
The third graph illustrates a very challenging case with extreme mismatch between training set (Caserta-WV-2) and target clips (Adelaide-WV-3) since even the sensor now is different.
As could be expected, the improvement achieved on Q8 through fine-tuning is always substantial, almost 0.2 on the average,
and significant improvements, although smaller, are observed also in terms of QNR.

These preliminary results are extremely encouraging.
In case of mismatch between training and target, the fine-tuning step improves significantly the objective measures of performance.
The impact on the quality of full resolution images, is more controversial.
This may be due to incorrect modeling of the MTF, but also to the limited ability of these measures to assess actual image quality.
In general, full-resolution quality assessment is an open issue \cite{Khan2009, Aiazzi2014, Palsson2016},
and the most sensible way to compare different solutions is to jointly look at the reduced-resolution and full-resolution results, and never neglect visual inspection.

\section{Experimental analysis}
\label{sec:exp}

To assess the performance of the proposed methods we carried out a number of experiments with real-world multiresolution images, exploring a wide range of situations.

In the following subsections we
\begin{itemize}
\item   list the methods under test, both proposed and reference;
\item   summarize the set of performance measures considered in the experiments, both full-reference and no-reference, discussing briefly their significance;
\item   describe training and test sets, and how they are combined to explore increasingly challenging cases;
\item   report and comment the experimental results, both numerical and visual;
\item   discuss the computational issue.
\end{itemize}

\subsection{Methods under analysis}

Our baseline is the PNN proposed in \cite{Masi2016}, with the input augmented by radiometric indexes.
However, since all the solutions proposed here have been implemented in Python using Theano \cite{Theano},
we have re-implemented in this framework also the baseline, originally developed with Caffe \cite{jia2014caffe}, in order to avoid biases due to the different arithmetic precision and/or randomization.
This justifies some small numerical differences w.r.t. results reported  in \cite{Masi2016}.
Furthermore, we consider three variations of PNN,
L1, obtained by replacing the L2 loss with L1 loss in PNN,
L1-RL, adopting in addition also the residual learning architecture, and
L1-RL-FT, which fine-tunes the network on the target.
All these methods share the same three-layer CNN, with hyper-parameters given in Tab.~\ref{tab:hyper}.
Only for the baseline, the input channels include also some radiometric indexes (red entries in Tab.~\ref{tab:hyper}).

\begin{table}
\centering
\setlength\tabcolsep{2.5pt}
\caption{CNN hyper-parameter for all proposed methods: \# output features $\times$ \# input channels $\times$ 2D filter support. Red entries only for baseline (input augmentation)}
\begin{tabular}{cccccc} \hline
\rule{0pt}{9pt} Sensor & ConvLayer 1                                    & Activ. & ConvLayer 2                    & Activ. & ConvLayer 3                   \\ \hline
\rule{0pt}{9pt} Ikonos & 48$\times$5/{\color{red} ~7}$\times$5$\times$5 & ReLU   & 32$\times$48$\times$5$\times$5 & ReLU   & 4$\times$32$\times$5$\times$5 \\
                GeoEye-1   & 48$\times$5/{\color{red} ~7}$\times$9$\times$9 & ReLU   & 32$\times$48$\times$5$\times$5 & ReLU   & 4$\times$32$\times$5$\times$5 \\
                WorldView-2   & 48$\times$9/{\color{red} 13}$\times$9$\times$9 & ReLU   & 32$\times$48$\times$5$\times$5 & ReLU   & 8$\times$32$\times$5$\times$5 \\
                WorldView-3   & 48$\times$9/{\color{red} 13}$\times$9$\times$9 & ReLU   & 32$\times$48$\times$5$\times$5 & ReLU   & 8$\times$32$\times$5$\times$5 \\ \hline
\end{tabular}
\label{tab:hyper}
\end{table}

Besides our CNN-based methods,
we consider a number of well-known conventional techniques selected because of their good performance,
and specifically,
PRACS \cite{Choi2011},
GSA \cite{Aiazzi2007},
Indusion \cite{Khan2008},
AWLP \cite{Otazu2005},
ATWT-M3 \cite{Ranchin2000},
MTF-GLP-HPM \cite{Aiazzi2003},
MTF-GLP-CBD \cite{Aiazzi2006},
BDSD \cite{Garzelli2008}, its recent extension
C-BDSD \cite{Garzelli2015},
and
SR \cite{Vicinanza2015},
based on sparse representations.
Details on these methods can be found in \cite{Vivone2015} and in \cite{Masi2016}, and obviously in the original papers.
In addition we also report as a naive reference the 23-tap polynomial interpolator (denoted EXP) used by many algorithms, including ours, as initial upsampler.
The software used to implement the methods and carry out all experiments is available online \cite{Grip} to ensure full reproducibility.

\subsection{Performance measures}

To assess performance we use the framework made available online \cite{OpenRS} by Vivone {\em et al.}, and described in \cite{Vivone2015}.
Accordingly, we report results in terms of multiple performance measures.
In fact, no single measure can be considered as a fully reliable indicator of pansharpening quality,
and it is therefore good practice to take into account different perspectives.
In particular, it is advisable to consider both full-reference (low-resolution) and full-resolution (no-reference) measures.

Following Wald's protocol \cite{Wald97,Ranchin2000,Vivone2015},
full-reference measures are computed on the reduced resolution dataset, so as to use the original MS data as reference.
Therefore, they can measure pansharpening accuracy {\em objectively}.
On the down side, the reduced-resolution data are obtained through a downgrading procedure which may introduce a bias in the accuracy evaluation.
A method that performs well on erroneously downgraded data may turn out to work poorly at full resolution.

The choice of the low-pass anti-aliasing filter which preceeds decimation is therefore a crucial issue of this approach.
In the original paper by Wald \cite{Wald97} it is left as an open concern.
Here,
we adopt the solution proposed by Aiazzi {\em et al.} \cite{Aiazzi2006}, and implemented in the Open Remote Sensing repository \cite{OpenRS, Vivone2015},
which uses a different low-pass filter for each MS band and for the PAN, matched to the specific channel MTF.

Here, we use the following widespread full-reference measures, referring to the original papers for their thorough description:
\begin{itemize}
\item {\bf SAM}: Spectral Angle Mapper \cite{Yuhas1992};
\item {\bf ERGAS}: Global adimens. relative synthesis error \cite{Wald2002};
\item {\bf Q}: Average universal image Quality index \cite{Zhou2002};
\item {\bf Q4 / Q8}: 4 / 8-band extension of {\bf Q} \cite{Alparone2004};
\end{itemize}

Full-resolution measures, instead, work on the original data, thus avoiding any biases introduced by the downgrading procedure.
In particular, we consider here the QNR and its components, referring again to the original paper for all details:
\begin{itemize}
\item {\bf QNR}: Quality with No-Reference index \cite{Alparone2008};
\item {\Dlambda}: Spectral component of QNR;
\item {\Desse}: Spatial component of QNR.
\end{itemize}
Their major drawback is, obviously, the absence of reference data at full resolution, which undermines the measures' objectivity.
In addition,
these measures rely on the upsampled MS image as a guide to compute intermediate quantities, thereby introducing their own biases.
In particular, for the EXP method, which performs only MS upsampling, the \Dlambda\ measure vanishes altogether,
leading to a very high QNR despite a clear loss of resolution and very poor results on reference-based measures.
More in general, \Desse\ is biased for any method approaching EXP.

In summary, the absence of true reference data makes assessment a challenging task.
Accordingly, a good pansharpening quality may be claimed when all measures, with and without reference, are good,
while results that change very much across measures suggest biases and poor quality.

{
\begin{table}
\centering
\renewcommand{\ru}{\rule{0mm}{3mm}}
\setlength\tabcolsep{3pt}
\caption{Test set and training set combinations used in the experiments.}
\begin{tabular}{lllcc} \hline
\ru Test dataset (\# clips) &  Training dataset   & Op. condition         & Table                 & Figure                         \\ \hline
\ru Caserta-IK (50)         &  Caserta-IK         & favourable            & \ref{tab:IK}          &                                \\
\ru Caserta-IK (50)         &  Caserta-GE-1       & challenging           & \ref{tab:IK}          &                                \\ \hline
\ru Caserta-GE-1 (70)       &  Caserta-GE-1       & favourable            & \ref{tab:GE}          &                                \\
\ru Caserta-GE-1 (70)       &  Caserta-IK         & challenging           & \ref{tab:GE}          &                                \\ \hline
\ru Caserta-WV-2 (30)       &  Caserta-WV-2       & favourable            & \ref{tab:WV}          &                                \\ \hline
\ru Stockholm-WV-2 (30)     &  Caserta-WV-2       & typical               & \ref{tab:Stockholm}   & \ref{fig:det1}, \ref{fig:det2} \\ \hline
\ru Washington-WV-2 (81)    &  Caserta-WV-2       & typical               & \ref{tab:Washington}  & \ref{fig:det1}, \ref{fig:det2} \\ \hline
\ru Adelaide-WV-3 (45)      &  Caserta-WV-2       & challenging           & \ref{tab:Adelaide}    & \ref{fig:det3}                 \\ \hline
\end{tabular}
\label{tab:datasets}
\end{table}
}

\subsection{Datasets and training}

To assess performance in a wide variety of operating conditions we experimented with images acquired by several multiresolution sensors,
Ikonos (IK for short), GeoEye-1 (GE-1), WorldView-2 (WV-2) and WorldView-3 (WV-3), with scenes concerning both rural and urban areas, taken in different countries.
Tab.~\ref{tab:datasets} reports the list of all test datasets.
Accordingly, the spatial resolution measured on the PAN component varies from 0.82m (IK) to 0.31m (WV-3),
while 4 (IK, GE-1) to 8 (WV) bands are available, covering the visible and near-infrared regions of the spectrum.

Our main goal, however, is to study performance as a function of the mismatch between training and test data.
Under this point of view, we classify operating conditions into favourable, typical, and challenging.
Favourable conditions occur when training and test sets, although separated, are taken from the same image, and hence share all major statistical features.
It is worth underlining that, due to lack of data, this is quite a common case, and the only one explored in our previous work \cite{Masi2016}.
More typycally, training and test data are expected to be unrelated.
For example, a network trained on the Caserta-WV-2 scene may be used for the pansharpening of the WV-2 image of Stockholm.
Finally, one may try to use a network trained on a given sensor to process images acquired from a different one.
For example, the network trained on the Caserta-WV-2 scene, may be used to pansharpen a WV-3 image.
In Tab.~\ref{tab:datasets}, together with each test dataset, we report the corresponding training set, with combinations covering all operating conditions of interest.
Note that only the Caserta datasets were used for training.
In particular,
for each sensor, a training/validation set was generated, comprising 14400/7200 tiles of 33$\times$33 pixels,
collected in mini-batches of 128 elements for an efficient implementation of the stochastic gradient descent algorithm.
The training procedure is the same already used in \cite{Masi2016} where additional details can be found.
Eventually, the trained nets are tested on a number of 1280$\times$1280 clips (PAN resolution), disjoint from the training set.

\subsection{Discussion of results}

\begin{table}
\caption{Performance indicators at reduced (full-reference) and full (no-reference) resolution on the Caserta-IK dataset.}
\centering
\image{TabIK}
\label{tab:IK}
\end{table}

\begin{table}
\caption{Performance indicators at reduced (full-reference) and full (no-reference) resolution on the Caserta-GE-1 dataset.}
\centering
\image{TabGE}
\label{tab:GE}
\end{table}

\begin{table}
\caption{Performance indicators at reduced (full-reference) and full (no-reference) resolution on the Caserta-WV-2 dataset.}
\centering
\image{TabWV}
\label{tab:WV}
\end{table}

We start our analysis from the {\em favourable} cases of Tables \ref{tab:IK}, \ref{tab:GE} and \ref{tab:WV} (discard the last two rows of Tables \ref{tab:IK}, \ref{tab:GE} for the time being),
where both training set and test set are drawn from the same image of Caserta, acquired by the IK, GE-1, and WV-2 sensors, respectively.
Results for conventional techniques are grouped in the upper part of the table, with the best result for each indicator shown in boldface blue.
Then, we have a single line for the baseline method, PNN with augmented input, implemented in Theano.
The following three rows show results for the proposed variations, with L1 loss, residual learning, and fine-tuning, with the best result in boldface red.
For these cases, it was already shown, in \cite{Masi2016}, that PNN improves significantly and almost uniformly over all measures with respect to reference methods.
Even the sparse-representation method, SR, which learns a dedicated dictionary 
for each target image,
is only competitive on no-reference indicators,
while showing large performance losses on full-reference ones.
Hence, we focus on the further modifications introduced in this work.
With respect to the baseline, L1 loss and residual learning guarantee small but consistent improvements, uniformly over all measures.
As for the fine-tuning step, it improves all full-reference measures but not all no-reference ones,
especially \Dlambda, probably due to the bias introduced in its computation.
In any case, the variant with fine-tuning is almost always the best CNN-based solution, and best overall, with some exceptions only for no-reference measures.
The limited improvement w.r.t. the baseline is explained by the very good results obtained by the latter in these favourable conditions.
It is worth reminding that PNN performs already much better than all conventional methods, and the gap has now grown even wider.
For the sake of brevity, we do not show visual results for these relatively less interesting cases.

Let us now consider the {\em typical} case, when training and test data are acquired with the same sensor but come from different scenes.
To this end we report in Tables \ref{tab:Stockholm} and \ref{tab:Washington} results for WV-2 data,
with network trained on the Caserta image and used on the Stockholm and Washington images.
In this case results are much more controversial.
PNN keeps providing good results, but not always superior to conventional methods,
especially PRACS, C-BDSD and SR on no-reference measures, and MTF-GLP-CBD 
on full-reference measures.
On Stockholm, in particular, a large \Dlambda\ value is observed, testifying of a poor spectral fidelity, which impacts on the overall QNR, 4 percent points worse than PRACS.
The adoption of L1 loss and residual learning provides mixed effects, mostly minor losses at low resolution and minor gains at full resolution.
On the contrary, fine-tuning has a strong impact on performance,
leading this version of PNN to achieve the best results almost uniformly on all measures and for both images.
For full reference measures, in particular, the fine-tuning version provides a huge gain with respect to both the baseline and the best conventional method.
On Stockholm, for example, SAM lowers from 7.45 to 4.82 and ERGAS from 5.58 to 3.72.
In terms of QNR it approaches very closely the best conventional methods,
while the best performance is given by naive interpolation (EXP), suggesting to take this indicator with some care.

\begin{figure*}
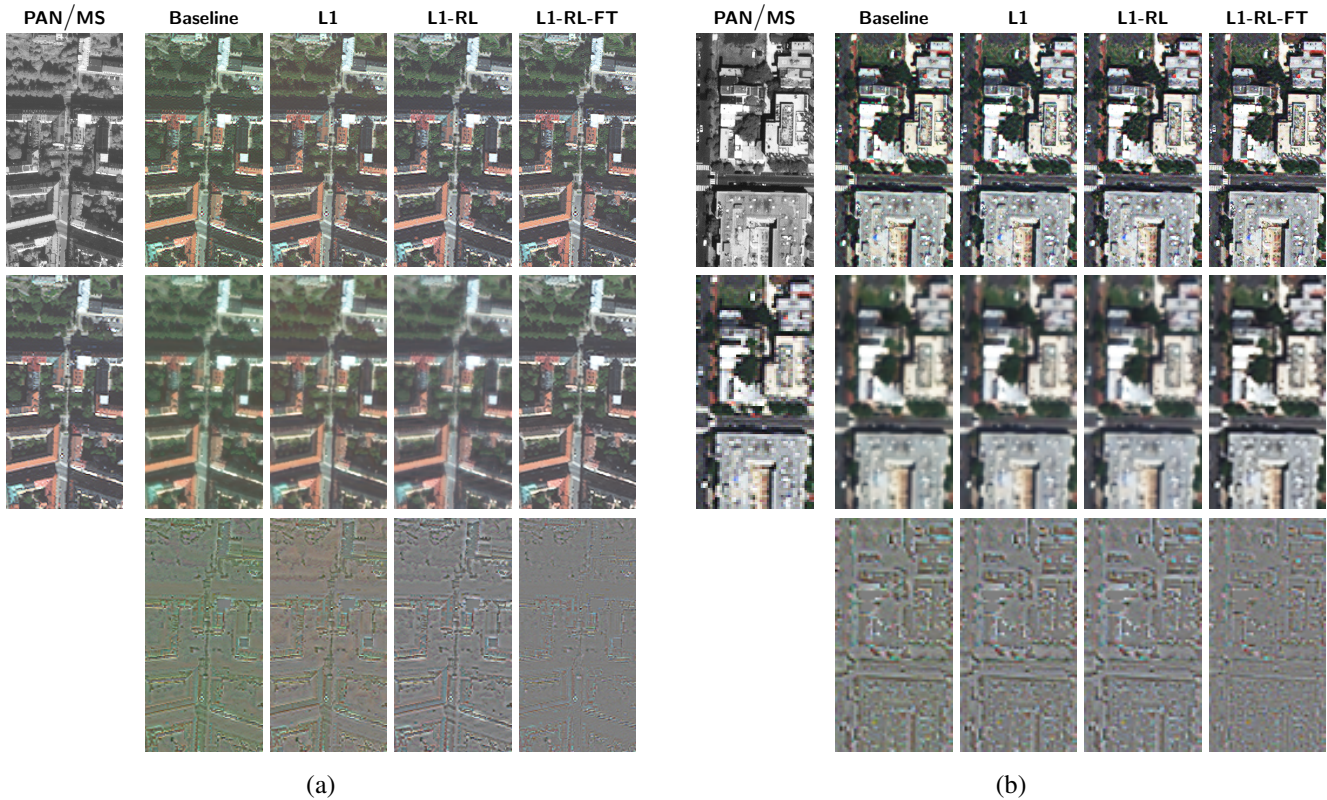

\centering
{
\setlength{\tabcolsep}{10pt}
\begin{tabular}{cc}
\image{T6Stock} & \image{T60Wash}\\
(a) & (b)
\end{tabular}
}
\caption{Output of baseline and proposed methods on clips from the Stockholm-WV-2 (a) and Washington-WV-2 (b) images.
From top to bottom: PAN $+$ results at full resolution, MS $+$ results at reduced resolution, error maps for the reduced-resolution case.}
\label{fig:det1}
\end{figure*}

Visual inspection helps explaining this behavior.
In Fig.~\ref{fig:det1} (a) and (b), with reference only to the proposed methods, we show some sample results for Stockholm and Washington, respectively.
Here and in the following figures,
we show in the first row the PAN image together with full-resolution pansharpening results, so as to appreciate spatial accuracy.
The second row, instead, shows the MS image followed by reduced-resolution pansharpening results, corresponding to full-reference measures, and providing information on spectral fidelity.
Then, in the third row, we show the difference between reduced-resolution pansharpening and MS, to better appreciate errors.
In each figure, all images are subject to the same histogram stretch to improve visibility, except for the difference images that are further enhanced.
The Stockholm image explains very clearly the relatively poor performance of PNN.
Because of the mismatch between training and test data, pansharpened images are affected by a large spectral error, both al low and high resolution, with a dominant green hue.
The problem is almost completely corrected by resorting to residual learning, which works on differences and, hence, tends to reduce biases.
However, there is still a clear loss of spatial resolution in the low-resolution image, as testified by structures in the error image.
Fine tuning solves also this problem, providing satisfactory results in terms of both spectral and spatial resolution.
The error image shows neither dominant hues (spectral errors) nor marked structures (spatial errors).
Similar considerations apply to the Washington image, even though in this case the spectral bias is much smaller, due to a better alignment of training and test data.

\begin{figure*}
\centering
\caption{Output of proposed and reference methods on clips from the Stockholm-WV-2 (top) and Washington-WV-2 (bottom) images.
PAN $+$ results at full resolution (1st and 4th rows), MS $+$ results at reduced resolution (2nd and 5th rows), error maps for the reduced-resolution case (3rd and 6th rows).}
\label{fig:det2}
\end{figure*}

In Fig.~\ref{fig:det2} the visual comparison is against reference methods.
The detail on top is particularly interesting because it includes an industrial plant with a large roof writing, the ideal image for visual appreciation of spatial resolution.
Let us first consider the EXP image (simple interpolation): despite the very strong blurring, it has the best no-reference measures,
further testifying on the need to consider multiple quality indicators, together with visual inspection.
At full resolution (first row) the proposed method is among the best, but some reference methods work equally well, notably MTF-GLP-CBD, Indusion, AWLP.
PRACS and ATWT-M3, instead, exhibit oversmoothing, while C-BDSD and ATWT-M3 are affected by spectral distortion.
Finally, SR provides sharp results but with some spatial artifacts which are clearly visible on the shadowed areas.
At reduced resolution, the proposed method seems clearly superior to all references, as predicted by objective measures and also confirmed by the error images in the third row.
Again, similar considerations with minor differences apply to the Washington image (bottom detail).

\begin{table}
\caption{Performance indicators at reduced (full-reference) and full (no-reference) resolution on the Stockholm-WV-2 dataset.}
\centering
\image{TabStock}
\label{tab:Stockholm}
\end{table}

\begin{table}
\caption{Performance indicators at reduced (full-reference) and full (no-reference) resolution on the Washington-WV-2 dataset.}
\centering
\image{TabWash}
\label{tab:Washington}
\end{table}

Finally, let us consider the most {\em challenging} case of cross-sensor pansharpening.
Going back to Tables \ref{tab:IK} and \ref{tab:GE}, in the last two rows we report results obtained with our best proposed method, L1-RL, using a cross-sensor network with and without fine-tuning.
In detail,
for Tab.~\ref{tab:IK}, concerning the Caserta-IK image, the network is trained on Caserta-GE-1 data, and viceversa for Tab.~\ref{tab:GE}.
In the absence of fine-tuning (penultimate row) there is a large loss of performance w.r.t same-sensor training, and even w.r.t. conventional methods, at least for full-reference measures.
This gap, however, is almost completely recovered through fine-tuning, achieving a performance which is only slightly inferior to the overall best.
It is worth reminding that this result is obtained with a negligible computational effort, while training a net from scratch would require many hours even with suitable GPUs.

\begin{figure*}
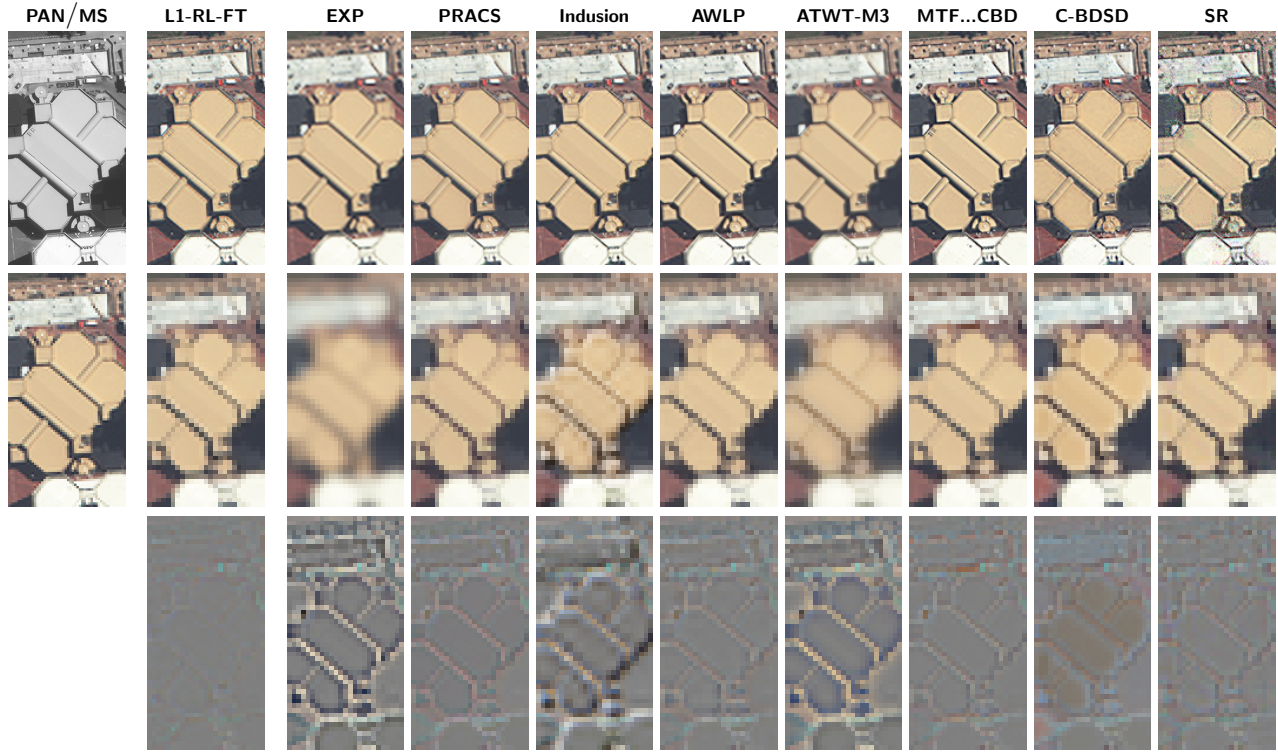

\centering
\image{T20bAdeC}
\caption{Output of proposed and reference methods on clips from the Adelaide-WV-3 image.
From top to bottom: PAN $+$ results at full resolution, MS $+$ results at reduced resolution, error maps for the reduced-resolution case.}
\label{fig:det3}
\end{figure*}

Tab.~\ref{tab:Adelaide} shows results for the Adelaide image, acquired with the WV-3 sensor, using the network trained on WV-2 data.
In this case there is no same-sensor comparison, we just used the available WV-2 net.
Again, after fine-tuning, the proposed method works much better than conventional methods for full-reference measures,
and only slightly worse than C-BDSD and SR for full-resolution measures.
The visual inspection of Fig.~\ref{fig:det3}, however,
shows that SR exhibits annoying spatial patterns at full-resolution, while C-BDSD is affected by spectral distortion.
The large performance gain w.r.t. all other methods appears also obvious from the analysis of error images,
since all reference methods present strong image-related structures (spatial distortion) and colored regions (spectral distortion).
Overall, thanks to fine-tuning, the proposed method seems preferable to all references even in a cross-sensor setting.

\begin{table}
\caption{Performance indicators at reduced (full-reference) and full (no-reference) resolution on the Adelaide-WV-3 dataset.}
\centering
\image{TabAde}
\label{tab:Adelaide}
\end{table}

Our analysis completes with some notes on computational complexity.
On a GPU-equipped computer (GeForce GTX Titan X, Maxwell, 12GB)
the proposed method runs very fast, taking little more than 1 second/Mpixel on the average, including fine-tuning.

Needless to say, this is the natural configuration for any method based on deep learning.
On the other hand, the algorithm can also run on a simple CPU-equipped computer, in which case complexity may become an issue.
Therefore, for all methods, we measured average CPU times (Intel Xeon E5-2670 1.80GHz, 64GB), reporting results in Tab.~\ref{tab:cplx}.
For 1280$\times$1280-pixel clips, running times go (excluding SR) from the 1.4 s/clip of Indusion and BDSD to the 17.3 s/clip of ATWT-M3.
Without fine-tuning, PNN-L1-RL is only somewhat slower, 24.4 s/clip, which is quite reasonable given the huge number of convolutions carried out in a CNN.
Of course, the iterative fine-tuning process, FT in the table, is much heavier, adding 210 seconds to the total CPU-time.
Whether this may prevent use of the proposed method depends only on user requirements.
On the other hand, SR suffers from the very same problem, due to on-line dictionary learning, and is even slower.

\begin{table}
\caption{CPU times (s) for $1280\times 1280$ and $2560\times 2560$-pixel clips.}
\centering
\image{TabCplx}
\label{tab:cplx}
\end{table}

Turning to 2560$\times$2560-pixel clips,
we notice that for almost all methods\footnote{We selected SR parameters to optimize performance, whatever the CPU-time.
Other choices are possible, but this goes beyond the scope of this work.}, including PNN,
CPU times scale about linearly with size.
However, the cost of fine-tuning does not increase,
since a good adaptation to the target image can be achieved based on a suitable subset of the image.
Therefore, its weight on the overall cost decreases,
becoming eventually negligible for the large-size images, say, 20000$\times$20000 pixels, used in real-world practice.

\section{Conclusions}
\label{sec:conclusion}

We started from our recently proposed \cite{Masi2016} CNN-based pansharpening method, featuring already a state-of-the-art performance,
and explored a number of architectural and operating variations to improve both quality and robustness.
When the training set is well matched to the test data,
residual learning and L1 loss ensure some limited improvements, together with a significant speed-up in training.
The most interesting results, however, are observed in the presence of training-test mismatch,
quite common in remote sensing due to the scarcity of free data.
In this case, target adaptation, obtained through a fine-tuning pass,
provides a very significant performance gain, with a reasonable computational cost and no active user involvement.

Full-resolution quality remains an open issue.
Indeed, while the performance is fully satisfactory on subsampled data, there is still room for improvements at the highest resolution.
Besides a better modeling of the MTF, and a better compensation of atmospheric effects \cite{Lolli2017},
a major impact may come from the design of more reliable no-reference measures.
These would enable training and fine-tuning based on a task-specific loss function, with a sure impact on performance.

\section{Acknowledgement}
We thank the Authors of all reference methods for sharing their codes, and those of the Open RS platform for providing the code for accuracy evaluation.
We also thank DigitalGlobe for providing the images used for training and testing our methods.

\balance

\bibliographystyle{IEEEtran}
{\footnotesize \bibliography{refs_compact}}

\end{document}